\documentclass[journal,article,submit,pdftex,moreauthors]
{Definitions/mdpi}
\let\linenumbers\relax
\firstpage{1}
\pubvolume{1}
\issuenum{1}
\articlenumber{0}
\pubyear{2026}
\copyrightyear{2026}
\datereceived{ }
\daterevised{ }
\dateaccepted{ }
\datepublished{ }

\usepackage{longtable}
\usepackage{caption}
\usepackage{algorithm}
\usepackage{algpseudocode}
\usepackage{listings}
\Title{ADAPT: Adaptive Decision and Abstention with Pool-Size-Aware Thresholds for Service Routing}

\Author{Manile Srun $^{1,}$*\orcidA{}, Muntasir Adnan $^{1}$\orcidB{}, Arslan Shaukat $^{1}$\orcidC{} and Carlos C. N. Kuhn $^{1}$\orcidD{}}
 
\address{%
$^{1}$ \quad Open Source Institute, Faculty of Science and Technology, University of Canberra, Canberra, Australia.}

\corres{Correspondence: manile.srun@canberra.edu.au}

\abstract{Retrieval-based classifiers efficiently route queries to specialised services, but they typically select a candidate even when no suitable service exists. 
We introduce Adaptive Decision and Abstention with
Pool-Size-Aware Thresholds for Service Routing (ADAPT), a lightweight routing architecture with an explicit fallback option that enables queries to abstain  from specialist selection. 
ADAPT uses the separation between the highest relevance score and the candidate score distribution as routing confidence. 
Experiments across retrieval models and increasing numbers of services reveal a consistent empirical behaviour: the confidence distribution changes systematically with candidate-pool size $K$, making fixed abstention thresholds increasingly miscalibrated. 
ADAPT addresses this effect with a $K$-aware threshold that adapts as services are added without classifier retraining. 
Evaluation across two specialist datasets shows that this simple mechanism provides an effective accuracy-cost trade-off compared with generative LLM routing, supporting lightweight and dynamically extensible service-routing architectures.}

\keyword{query routing; retrieval models; large language models; out-of-distribution detection; threshold calibration; multi-service systems}

\begin{document}

\section{Introduction}

Modern AI systems increasingly orchestrate collections of specialised models, tools, knowledge sources and external services rather than relying on a single general-purpose model
~\cite{adnan2024unleashing, su2025toolorchestra, shen2023hugginggpt}.
Such architectures require a routing mechanism that maps each incoming query to the capability best suited to process it. This pattern appears in model routing, tool-augmented language models, multi-agent systems and modular AI architectures.
In systems such as the Cognitive System of Machine Intelligent Computing (CoSMIC) ~\cite{adnan2024unleashing}, for example, a query analyser sits between the user and heterogeneous specialist services such as a chess engine and a vector-database update service, and selects the capability that should process each request. 
This architectural pattern lets you add and update capabilities independently, but it also makes routing harder as the number and diversity of available specialists grow~\cite{sorathiya2026enrich}. 
A practical router must therefore select the appropriate specialist accurately, introduce little computational and latency overhead, remain effective as the candidate pool grows~\cite{ratnakar2026latentgate,sorathiya2026enrich}, and accommodate new or updated specialists without requiring repeated retraining.

Several routing strategies have been developed across related settings. Model-routing approaches select among candidate LLMs according to expected quality, cost, or task suitability, whereas capability-routing approaches select among heterogeneous specialised models, tools, or services. Researchers have used both generative models and learned routers for these decisions.
For example, CoSMIC~\cite{adnan2024unleashing} uses an LLM-based query analyser, while RouteLLM~\cite{ong2025routellm}, HybridLLM~\cite{ding2024hybridllm}, RouterDC~\cite{chen2024routerdc}, FrugalGPT~\cite{chen2024frugalgpt}, AutoMix~\cite{aggarwal2024automix}, MoDEM~\cite{simonds2024modem}, ToolOrchestra~\cite{su2025toolorchestra} and LatentGate~\cite{ratnakar2026latentgate} rely on learned routing mechanisms. 
These approaches show the growing importance of routing, but generative routers add the cost of a full model call to every request, while learned routers must be retrained whenever specialists change.
More recently, retrieval has emerged as a lightweight alternative for selecting tools or capabilities. 
They can be deployed without task-specific router training by matching a query against textual descriptions of the available capabilities, as in ToolRet~\cite{shi2025toolret}, Re-Invoke~\cite{chen2024reinvoke}, Tool-DE~\cite{lu2025toolde}, Tool-to-Agent Retrieval~\cite{lumer2025tooltoagent}, mmRAG~\cite{xu2025mmrag} and Enrich-Retrieve-Rank~\cite{sorathiya2026enrich}. 
These studies indicate that retrieval can provide an efficient and extensible mechanism for reducing the search space of increasingly large capability collections. 
Nevertheless, retrieval alone is not yet a complete router. 
A retrieval model ranks candidates but does not decide whether any are appropriate, so it returns the highest-scoring specialist even for a query that falls outside every specialist's capabilities.
A practical router needs a \emph{fallback} mechanism that sends such queries to a general service when the top specialist is not convincing enough, and this mechanism must stay reliable as specialists are added.
How well such a router works also depends on the retrieval architecture, from bi-encoders~\cite{reimers2019sbert} and late-interaction models such as ColBERT~\cite{khattab2020colbert} to cross-encoders~\cite{nogueira2019passage}, which differ in accuracy and cost, and it remains unclear how these families compare with LLM routers.

In this work, we introduce Adaptive Decision and Abstention with Pool-Size-Aware Thresholds for Service Routing (ADAPT), a lightweight retrieval-based routing architecture that extends a conventional classifier with an explicit fallback option.
Rather than relying on the top specialist's absolute score, ADAPT measures how far that score stands out from the median score of all specialists. 
This separation acts as a confidence signal, calibrated on off-topic queries alone, and queries with a gap that is too small fall back to the general service.
Because specialists are represented by descriptions rather than fixed classifier outputs, we can add or update them without retraining the retrieval model. 
We evaluate ADAPT with bi-encoders, ColBERT models and cross-encoders against LLM routers, on two specialist domains and pools of up to 14 specialists.

This work makes two contributions.
\begin{itemize}
\item We identify an empirical property of retrieval-based routing. 
The confidence required to distinguish specialist and off-topic queries changes with the number of available specialists, because off-topic score separation grows with $K$ across all retrieval architectures we test. 
A threshold calibrated for a small candidate pool therefore becomes increasingly permissive as specialists are added.
\item We introduce ADAPT, in which the routing threshold is modelled as a function of $K$ and fitted per model using off-topic queries alone.
We show that this pool-size-aware calibration reduces the degradation in off-topic rejection as the service collection grows, and that ADAPT rivals LLM routers at a fraction of their cost.
\end{itemize}

Although we observe similar growth trends across retrieval architectures, we use this relationship as an empirical basis for adapting the routing threshold, without assuming that it generalises universally. 
The observed convergence after score normalisation suggests potentially model-agnostic scaling behaviour, whose statistical and theoretical foundations remain open to future investigation.

\section{Related Work}

Language model routing aims to select the most suitable model or service for a given query, balancing task performance, computational cost, and resource requirements. 
RouteLLM~\cite{ong2025routellm}, HybridLLM~\cite{ding2024hybridllm} and RouterDC~\cite{chen2024routerdc} train routers, where RouteLLM and HybridLLM send easy queries to a smaller model and RouterDC sends each query to the LLM expected to answer it best.
FrugalGPT~\cite{chen2024frugalgpt} and AutoMix~\cite{aggarwal2024automix} escalate to a larger model only when needed. 
Other methods route each request to a specialised model or service.
MoDEM~\cite{simonds2024modem} fine-tunes a domain classifier, CoSMIC~\cite{adnan2024unleashing} uses an LLM-based query analyser, ToolOrchestra~\cite{su2025toolorchestra} trains a small orchestrator with reinforcement learning, and LatentGate~\cite{ratnakar2026latentgate} trains a lightweight classifier over a frozen language model.

Retrieval models are widely used to select tools or capabilities for LLM agents. 
ToolRet~\cite{shi2025toolret} compares keyword-based and semantic retrieval for selecting tools from large tool collections.
Re-Invoke~\cite{chen2024reinvoke} and Tool-DE~\cite{lu2025toolde} improve retrieval by enriching tool descriptions.
Tool-to-Agent Retrieval~\cite{lumer2025tooltoagent} selects agents using the capabilities of their tools, not only broad agent descriptions.
mmRAG~\cite{xu2025mmrag} evaluates query routing across heterogeneous data sources, with a baseline that matches queries to dataset descriptions using dense embeddings.
Enrich-Retrieve-Rank~\cite{sorathiya2026enrich} shortlists capabilities with keyword-based and embedding-based retrieval, and an LLM then ranks them.

Abstention lets a model decline to answer instead of risking an error, and it has long been studied for classification~\cite{chow1957,elyaniv2010}.
It is closely related to detecting out-of-distribution inputs~\cite{liang2017odin} and prediction errors~\cite{hendrycks2016baseline}.
In retrieval, it has received little attention~\cite{gisserotboukhlef2024trustworthy}. 
Retrieval systems return a ranking even when the collection holds no relevant item, so a few recent works study when a system should abstain, either from the relevance scores alone, such as the maximum score or the gap between the top two scores~\cite{gisserotboukhlef2024trustworthy}, or with a classifier trained on in-domain and out-of-domain queries~\cite{triantafyllopoulos2026knowing}.

Taken together, these works leave three gaps. 
Most routing studies compare learned and LLM-based routers, but not retrieval models of different architectures against LLM routers on identical queries, especially as the specialist pool grows. 
Additionally, retrieval-based capability selection typically assumes that a suitable capability exists, so it cannot send an off-topic request elsewhere. 
Current retrieval abstention methods are developed for a single ranker over a single knowledge base and do not consider how confidence behaves as the number of candidate specialists grows.
We investigate this effect by comparing retrieval-based and LLM-based routers over specialist pools ranging from $K=3$ to $K=14$. Based on the observed behaviour of off-topic queries, we introduce a pool-size-aware threshold that adapts the fallback decision as the number of available specialists changes.

\section{Methodology}
\label{sec:method}

\subsection{Problem formulation}
\label{sec:problem}

Let there be $K \ge 3$ specialist services, numbered $1,\dots,K$, each covering one topic; and one general service that acts as a fallback.
A query $q$ is \emph{in-scope} if it belongs to the topic of exactly one specialist, and \emph{off-topic} if it belongs to none.
For each query, let $s_i(q)$ be the relevance score that a pre-trained scoring model assigns to specialist $i$.
The router $r$ returns the specialist service that handles $q$, namely the top-scoring specialist if the gap $g(q)$ (Section~\ref{sec:gap}) exceeds a cutoff $T(K)$ (Section~\ref{sec:cutoff}), and the general service otherwise:
\begin{equation}
  r(q)=
  \begin{cases}
    \arg\max_{i} s_i(q) & \text{if } g(q)>T(K),\\
    \text{general} & \text{otherwise.}
  \end{cases}
  \label{eq:router}
\end{equation}

\subsection{Gap-based decision rule}
\label{sec:gap}

\begin{figure}[H]
  \centering
  \includegraphics[width=\linewidth]{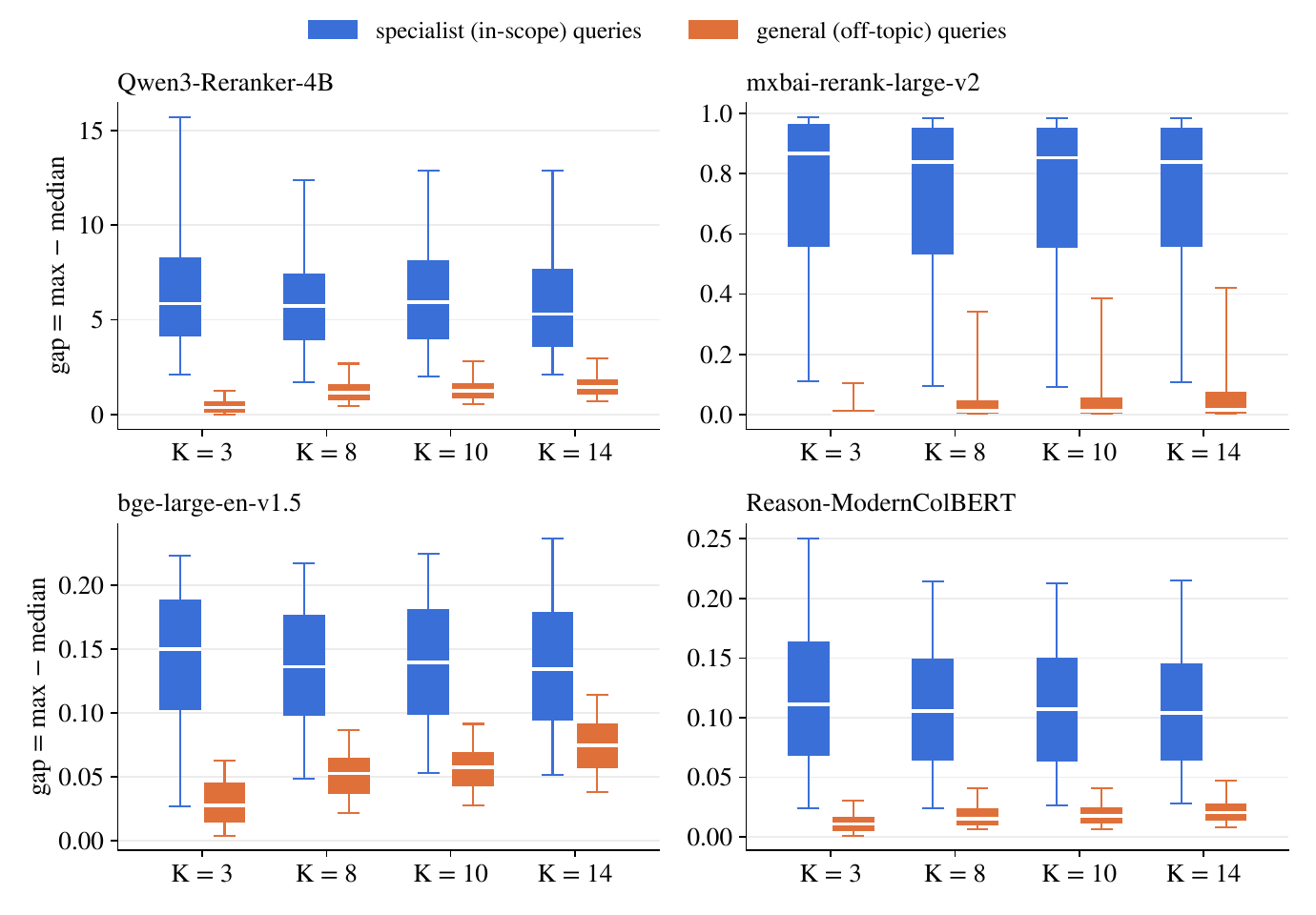}
  \caption{Gap between the highest and the median specialist score for in-scope (blue, 100 MMLU test queries per specialist) and off-topic (orange, 300 CLINC-OOS and BANKING77 queries) test queries at $K = 3$, $8$, $10$ and $14$, before any gate is applied. The four representative models are two cross-encoders (Qwen3-Reranker-4B, mxbai-rerank-large-v2), a bi-encoder (bge-large-en-v1.5) and a ColBERT model (Reason-ModernColBERT).}
  \label{fig:gap_boxplots}
\end{figure}

The decision rule is inspired by an observation on the scores, shown in
Figure~\ref{fig:gap_boxplots}. 
For in-scope queries, the highest score lies well above the typical (median) score, whereas for off-topic queries the scores are generally similar, so the difference is small. 

We therefore define the \emph{gap} of a query as the difference between its highest and its median score,
\begin{equation}
  g(q)=\max\nolimits_{i} s_i(q)-\operatorname{median}_{i} s_i(q)
  \label{eq:gap}
\end{equation}
which measures how far the best specialist stands out from the other scores and helps to separate the two types of query.

\subsection{Empirical Pool-Size Scaling of the Abstention Threshold}
\label{sec:cutoff}

The router compares a query's gap with a threshold $T(K)$ that depends on the number of specialists. 
We observed that the off-topic gaps increase with $K$ (Figure~\ref{fig:loglaw_fit}). 
We model this dependence with the two-parameter function
\begin{equation}
    T(K)=c_0+c_1\ln K
    \label{eq:cutoff}
\end{equation}
where $c_0$ and $c_1$ are model-specific calibration constants. 
We selected the logarithmic form as a parsimonious description of the observed behaviour. 
The threshold increases with candidate-pool size, but the incremental adjustment required for each additional specialist diminishes as the pool grows. 
This behaviour is captured by 
\begin{equation} 
    \frac{\mathrm{d}T(K)}{\mathrm{d}K} = \frac{c_1}{K}
    \label{eq:cutoff_derivative} 
\end{equation} 
so that, for $c_1>0$, the threshold increases monotonically while its marginal rate of increase decreases with $K$. 
This provides a simple pool-size correction using only two calibration parameters, while remaining consistent with the empirical cutoff curves observed across the retrieval models evaluated in this study. 
For each scoring model, we estimate $c_0$ and $c_1$ by least-squares fitting (L2 loss) to the 90th-percentile off-topic gaps obtained for $K=3,\ldots,8$ (Section~\ref{sec:fitting}). 
We do not use candidate-pool sizes above $K=8$ during fitting; therefore, they provide an extrapolative evaluation of Equation~\eqref{eq:cutoff}.

\subsubsection{Fitting the constants}
\label{sec:fitting}

\begin{figure}[H]
  \centering
  \includegraphics[width=\linewidth]{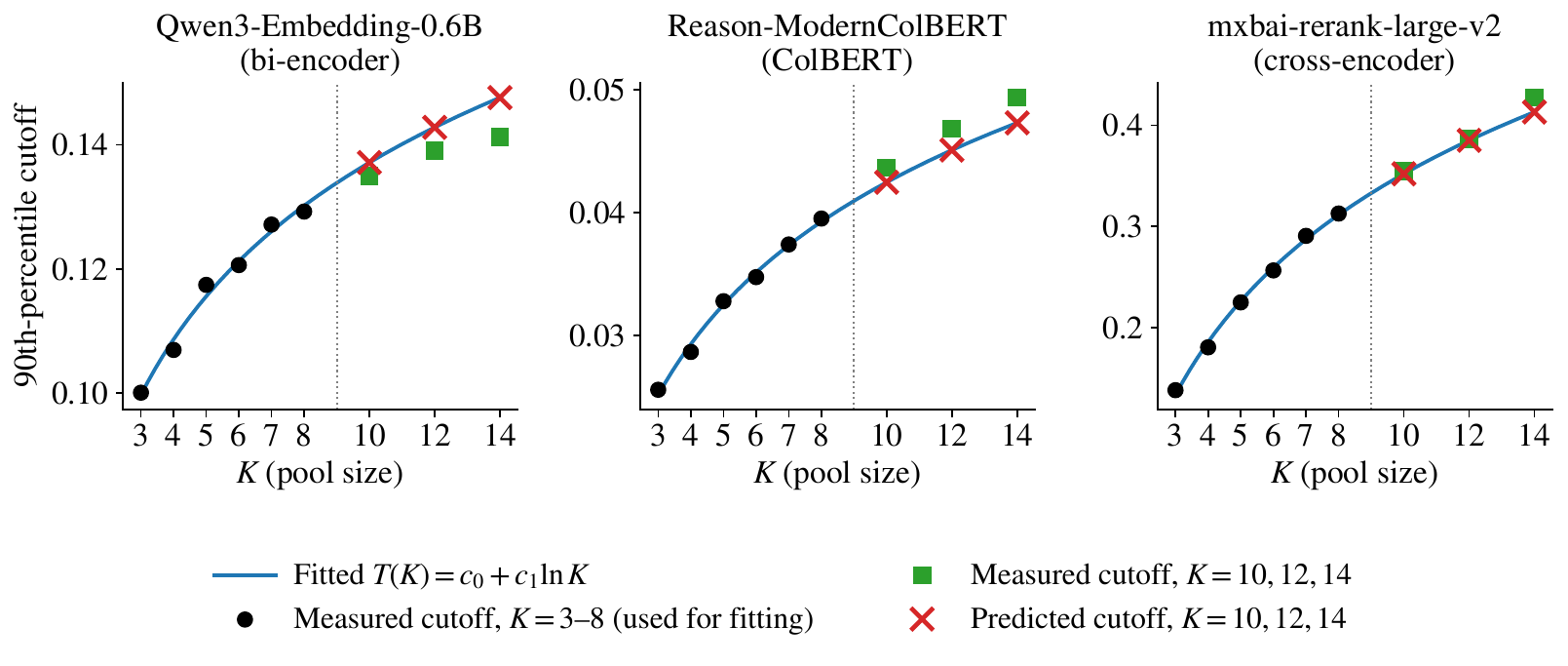}
  \caption{Fitted $T(K)=c_0+c_1\ln K$ for one example model from each family,
  fitted on $K=3,\dots,8$ and tested on $K=10$, $12$ and $14$.}
  \label{fig:loglaw_fit}
\end{figure}

For each scoring model, we fit $c_0$ and $c_1$ in Equation~\eqref{eq:cutoff} on the 100 off-topic queries of the CLINC-OOS~\cite{larson-etal-2019-evaluation} validation split, using the pool of 14 specialists from MMLU~\cite{hendryckstest2021}. 
Thus, for each calibration pool size $K$, the threshold estimate exhaustively aggregates all $\binom{14}{K}$ possible specialist subsets rather than depending on a particular ordering or nested pool. 
For each group, we compute the gap for every query (Equation~\eqref{eq:gap}), take the 90th percentile of the 100 gaps (the choice of $p=90$ is examined in Appendix~\ref{app:percentile}), and average over all groups of the same size to get one cutoff per $K$. 
The constants are the least-squares fit of $T(K)$ to these six cutoffs. 
We compute the cutoffs for $K=10$, $12$, and $14$ the same way but do not use them in the fit; we use them to evaluate the extrapolation (Figure~\ref{fig:loglaw_fit}). 
All cutoffs are computed from the off-topic validation queries only, so no labelled in-scope data is needed, and the test queries play no part in the fit.

To compare all 14 retrieval models on one axis, we rescale the measured cutoffs $\bar{T}_m(K)$ of each model $m$ as $(\bar{T}_m(K)-\bar{T}_m(3))/c_{1,m}$, since their sizes differ by several orders of magnitude.
\begin{figure}[H]
  \centering
  \includegraphics[width=0.5\linewidth]{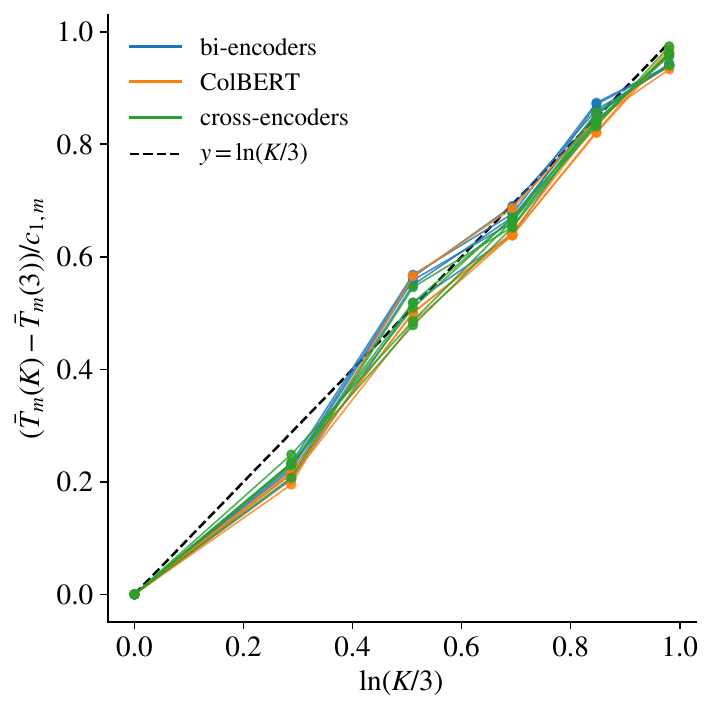}
  \caption{Rescaled measured cutoffs of the 14 retrieval models for
  $K=3,\dots,8$, coloured by family. The dashed line marks exactly
  logarithmic cutoffs.}
  \label{fig:collapse}
\end{figure} 
If the measured cutoffs follow Equation~\eqref{eq:cutoff} exactly, the rescaled values equal $\ln(K/3)$ for every model, and Figure~\ref{fig:collapse} shows that all models lie close to this line.
The cutoffs of all models therefore approximately follow a logarithmic shape over $K=3,\dots,8$, and the models differ only in the constants $c_{0,m}$ and $c_{1,m}$.

\subsection{Use of AI Tools in Manuscript Preparation}
\label{sec:gen_ai_usage}

Generative AI tools (Claude Sonnet 5.5, Anthropic; GitHub Copilot with GPT-5.6, OpenAI) were used during manuscript preparation for two purposes. 
First, they assisted in drafting and refining explanatory text, including identifying inconsistencies between reported claims and their supporting data. 
Second, they assisted in debugging during implementation and with the plotting code used to prepare the figures.

\section{Experimental Setup}
\label{sec:setup}

All experiments ran on a machine with two NVIDIA A100 80GB (PCIe) GPUs.

\subsection{Dataset}
\label{sec:dataset_setup}
We use three datasets. 
MMLU~\cite{hendryckstest2021} provides the specialists, and BANKING77~\cite{casanueva2020banking77} and CLINC-OOS~\cite{larson-etal-2019-evaluation} provide the off-topic queries. 
MMLU consists of multiple-choice questions from 57 subjects. 
We treat each subject as a specialist. 
Its validation split is used to build the specialist descriptions (Section~\ref{sec:specialist_description}) and for the percentile analysis (Appendix~\ref{app:percentile}), and its test split is used for testing, with 100 queries per specialist. 
BANKING77 contains 13,083 customer-service queries from the banking domain annotated with 77 intents. 
CLINC-OOS has 23,700 queries, made up of 22,500 in-scope queries across 150 intents and 10 domains and 1,200 out-of-scope queries.
We fit the threshold on 100 queries from the CLINC-OOS validation split (Section~\ref{sec:fitting}). 
For testing, we combine 150 queries from the BANKING77 test split and 150 queries from the CLINC-OOS test split into 300 off-topic queries.

\subsection{Constructing Specialists and Descriptions}
\label{sec:specialist_description}

Each specialist is one MMLU subject. 
We test pool sizes $K = 3, 5, 8, 10, 12$ and $14$. 
The pools are nested, so each pool contains the previous one, and Table~\ref{tab:specialists} lists the specialists added at each size.

\begin{table}[H]
\caption{Specialists in each pool. Each pool adds specialists to the previous one. Every specialist has 100 test queries, and the same 300 off-topic queries are used for every pool.}
\label{tab:specialists}
\begin{tabularx}{\textwidth}{cX}
\toprule
\textbf{Pool size} & \textbf{Specialists}\\
\midrule
3  & Abstract algebra, anatomy, astronomy\\
5  & Pool of 3 plus business ethics, college computer science\\
8  & Pool of 5 plus high school chemistry, high school geography, jurisprudence\\
10 & Pool of 8 plus world religions, formal logic\\
12 & Pool of 10 plus marketing, sociology\\
14 & Pool of 12 plus high school psychology, prehistory\\
\bottomrule
\end{tabularx}
\end{table}
Each specialist is represented by a short text description that the models use to choose it. 
We set up the descriptions from all questions in the MMLU validation split for each specialist (between 11 and 60 per specialist), which are not used for testing. 
We build the descriptions in three steps.
\begin{equation}
\text{Validation Questions} \rightarrow \text{Extracted Concepts} \rightarrow \text{Core Concepts} \rightarrow \text{Description}
\label{eq:description_pipeline}
\end{equation}
First, Qwen2.5-32B reads the subject's validation questions and answer choices and extracts the concepts they test, giving a fine-grained list of 9 to 58 concepts depending on the number of validation questions.
Second, the same model selects the 6 to 9 most representative concepts, reducing near-duplicates and giving each specialist a similar number of concepts.
Third, Qwen3-32B writes a 40- to 60-word description from the core concepts and the full set of extracted concepts.
All steps run with temperature 0 in Ollama~\cite{ollama}. 
For a quick setup of the descriptions, no model reviews another model's output, and we use two different models to reduce self-confirmation by a single model.

\subsection{Routing Models}

We compare bi-encoders, ColBERT models and cross-encoders (Sentence
Transformers~\cite{reimers2019sbert}) with LLM routers (Ollama~\cite{ollama}).

\textbf{Bi-encoders.} A bi-encoder embeds the query and each description separately and scores them by cosine similarity. 
We compute the description vectors once, so a query costs one encoder pass. We use bge-large-en-v1.5~\cite{xiao2023bge}, all-MiniLM-L6-v2~\cite{{wang2020minilm, reimers2019sbert}}, and Qwen3-Embedding at 0.6B, 4B and 8B~\cite{zhang2025qwen3embedding}.
\begin{algorithm}[H]
\caption{Bi-encoder scoring}
\label{alg:biencoder}
\begin{algorithmic}[1]
\Require query $q$, descriptions $d_1,\dots,d_K$, bi-encoder $f$
\Ensure scores $s_1(q),\dots,s_K(q)$
\State $\mathbf{v}_i \gets f(d_i)$ for $i=1,\dots,K$ \Comment{one vector, once}
\State $\mathbf{u} \gets f(q)$ \Comment{one vector}
\State $s_i(q) \gets \operatorname{cosinesim}(\mathbf{u}, \mathbf{v}_i)$ for $i=1,\dots,K$
\end{algorithmic}
\end{algorithm}

\textbf{Cross-encoders.} A cross-encoder reads the query and one description together in a single pass and outputs one relevance score. 
We use bge-reranker-v2-m3~\cite{chen2024bge, li2023making}, Qwen3-Reranker at 0.6B, 4B and 8B~\cite{zhang2025qwen3embedding}, and mxbai-rerank-large-v2~\cite{lee2025mxbairerank, v2rerank2025mxbai}.
\begin{algorithm}[H]
\caption{Cross-encoder scoring}
\label{alg:crossencoder}
\begin{algorithmic}[1]
\Require query $q$, descriptions $d_1,\dots,d_K$, cross-encoder $f$
\Ensure scores $s_1(q),\dots,s_K(q)$
\State $s_i(q) \gets f(q, d_i)$ for $i=1,\dots,K$ \Comment{one joint pass per pair}
\end{algorithmic}
\end{algorithm}

\textbf{ColBERT models.} A ColBERT model~\cite{khattab2020colbert} keeps one vector for each token and not one for the whole text. 
Each query token is matched to its most similar description token, and these best matches are summed (MaxSim). 
Longer queries give larger sums, so we divide each score by $n$, the number of tokens in the query. 
We use ColBERTv2~\cite{santhanam2022colbertv2}, mxbai-colbert-large-v1~\cite{mxbai-colbert}, Reason-ModernColBERT~\cite{reasonmoderncolbert} and LFM2-ColBERT-350M~\cite{lfm2colbert}.
\begin{algorithm}[H]
\caption{ColBERT scoring}
\label{alg:colbert}
\begin{algorithmic}[1]
\Require query $q$, descriptions $d_1,\dots,d_K$, ColBERT encoder $f$
\Ensure scores $s_1(q),\dots,s_K(q)$
\State $\mathbf{v}_i \gets f(d_i)$ for $i=1,\dots,K$ \Comment{one vector per token, once}
\State $\mathbf{u} \gets f(q)$ \Comment{one vector per token, $n$ tokens}
\State $s_i(q) \gets \mathrm{MaxSim}(\mathbf{u}, \mathbf{v}_i)\,/\,n$ for $i=1,\dots,K$
\end{algorithmic}
\end{algorithm}

\textbf{LLM routers.} An LLM router receives the descriptions of all specialists and the query in one prompt (Appendix~\ref{app:router_prompt}). 
It answers with the name of one specialist, or with $-1$ when no description fits, which sends the query to the general service. 
It decides by itself, so it produces no scores and the gate is not used. 
An answer that is neither the name of a specialist nor the general service ($-1$) counts as incorrect. We test Llama~3.1~8B~\cite{grattafiori2024llama3}, Phi-4~14B~\cite{abdin2024phi4}, Qwen2.5 at 7B, 14B and 32B~\cite{qwen2025qwen25}, and Qwen3~32B~\cite{yang2025qwen3}, served with Ollama~\cite{ollama}. 

\subsection{Metrics}

We report three numbers.
\emph{Specialist recall} is the percentage of specialist queries that the model sends to the correct specialist.
A specialist query sent to another specialist or to the general service counts as a miss.
\emph{General recall} is the percentage of off-topic queries that the model sends to the general service.
\emph{Accuracy} is the percentage of all test queries that the model sends to the expected service.
For the cost, we report the energy needed to route the test queries, measured with CodeCarbon~\cite{codecarbon}.

\subsection{Second Domain Setup}
\label{sec:MEDMCQA_setup}

To test the gate outside MMLU, we use MedMCQA~\cite{pal2022medmcqa}, a multiple-choice dataset of medical questions grouped by specialty. 
We use nine specialties as specialists: \emph{Radiology, Social \& Preventive Medicine, Forensic Medicine, Ophthalmology, Microbiology, Gynaecology \& Obstetrics, ENT, Skin} and \emph{Dental}. 
Their descriptions are built as in Section~\ref{sec:specialist_description}, and each specialty has 60 test questions. 
As with MMLU, we use pools of $K=3$, $5$, $7$ and $9$ specialists, where each pool is nested in the next.
The general queries are from four MMLU subjects: \textit{high school US history, abstract algebra, business ethics and marketing}. 
We use 75 questions from each, so 300 general queries for every $K$.

\section{Results}
\label{sec:results}

\subsection{Can retrieval models route with a fallback mechanism?}
\label{sec:res_accuracy}

Table~\ref{tab:acc-k} shows two main things. 
First, \textbf{retrieval models can provide competitive routing performance without generative inference in the evaluated settings}. 
A model that only scores how well a query fits each description can route as well as an LLM when the pool is small, and stays competitive when the pool is large. 
For example, Qwen3-Reranker-4B, which has eight times fewer parameters than qwen2.5:32b, is more accurate at every $K$ (97.50\% against 88.67\% at $K=3$ and 81.35\% against 79.76\% at $K=14$), and for $K\leq 5$ it is more accurate than all six LLMs. 
For $K\geq 8$ the most accurate model is the LLM qwen3:32b, but Qwen3-Reranker-8B trails it by only 2.5 to 4.4 points.
Even small retrieval models such as all-MiniLM-L6-v2 (about 22 million parameters) and Qwen3-Reranker-0.6B are more accurate than the two LLMs with 7 to 8 billion parameters at every $K$.
Retrieval models come close to the best LLM's accuracy \textit{without generating any text, which saves all output tokens}. 
Second, \textbf{a larger pool of services lowers the accuracy of every family, but not by the same amount}.
Taking one representative from each family, we observe that from $K=3$ to $K=14$ the bi-encoder bge-large-en-v1.5 loses the most (22.7 points), followed by the ColBERT model Reason-ModernColBERT (19.3), while the cross-encoder Qwen3-Reranker-4B loses the least (16.2) and still stays above qwen2.5:14b, whose accuracy falls by 19.2 points. 
The only exception is phi4:14b, whose accuracy rises until $K=8$ and then falls like the others, because with few services it often adds an explanation to its answer, which we count as wrong. 
This shows that LLM routers can return invalid answers when they do not follow the output instructions, a failure that retrieval models cannot make, since they always select one of the listed services.

\begin{table}[H]
\caption{Routing accuracy (\%) of each model for each number of specialist services $K$ with fallback. 
Bold marks the best retrieval model and the best LLM in each column.
\label{tab:acc-k}}
\centering
\resizebox{\linewidth}{!}{%
\begin{tabular}{llrrrrrr}
\toprule
\textbf{Family} & \textbf{Router} & $K=3$ & $K=5$ & $K=8$ & $K=10$ & $K=12$ & $K=14$ \\
\midrule
Bi-encoder & bge-large-en-v1.5    & 91.67 & 89.00 & 82.82 & 80.77 & 71.27 & 69.00 \\
           & Qwen3-Embedding-0.6B & 83.00 & 78.75 & 71.27 & 69.77 & 64.67 & 62.18 \\
           & Qwen3-Embedding-4B   & 89.67 & 75.12 & 65.64 & 64.46 & 57.53 & 54.76 \\
           & Qwen3-Embedding-8B   & 88.67 & 83.00 & 74.55 & 71.69 & 66.13 & 63.06 \\
           & all-MiniLM-L6-v2     & 88.83 & 80.50 & 74.73 & 73.62 & 69.53 & 66.82 \\
\midrule
ColBERT    & LFM2-ColBERT-350M      & 93.33 & 85.12 & 81.45 & 77.38 & 71.60 & 69.59 \\
           & colbertv2.0            & 77.00 & 66.12 & 58.36 & 57.46 & 54.67 & 50.47 \\
           & Reason-ModernColBERT   & 91.67 & 89.88 & 83.55 & 80.46 & 74.87 & 72.35 \\
           & mxbai-colbert-large-v1 & 80.50 & 71.12 & 67.91 & 64.00 & 60.33 & 56.94 \\
\midrule
Cross-encoder & bge-reranker-v2-m3    & 92.83 & 84.50 & 72.00 & 70.23 & 65.00 & 59.71 \\
              & Qwen3-Reranker-0.6B   & 94.17 & 87.50 & 82.36 & 78.62 & 73.53 & 70.35 \\
              & Qwen3-Reranker-4B     & \textbf{97.50} & \textbf{95.12} & 87.64 & 87.15 & 82.73 & 81.35 \\
              & Qwen3-Reranker-8B     & 87.50 & 93.00 & \textbf{90.09} & \textbf{89.85} & \textbf{84.80} & \textbf{82.47} \\
              & mxbai-rerank-large-v2 & 94.67 & 90.00 & 84.09 & 82.54 & 76.73 & 74.71 \\
\midrule
LLM & qwen2.5:7b  & 74.33 & 64.12 & 62.91 & 57.62 & 58.13 & 54.88 \\
    & llama3.1:8b & 77.67 & 69.88 & 73.91 & 70.31 & 68.07 & 66.18 \\
    & phi4:14b    & 71.17 & 79.12 & 86.64 & 86.08 & 81.60 & 79.47 \\
    & qwen2.5:14b & \textbf{96.67} & 92.88 & 84.27 & 85.62 & 78.47 & 77.47 \\
    & qwen2.5:32b & 88.67 & 86.62 & 84.36 & 85.77 & 81.53 & 79.76 \\
    & qwen3:32b   & 96.33 & \textbf{94.12} & \textbf{92.64} & \textbf{92.31} & \textbf{87.60} & \textbf{86.88} \\
\bottomrule
\end{tabular}}
\end{table}

\subsection{How well are specialist and off-topic queries routed?}
\label{sec:res_recall}

\begin{figure}[H]
\centering
\includegraphics[width=\linewidth]{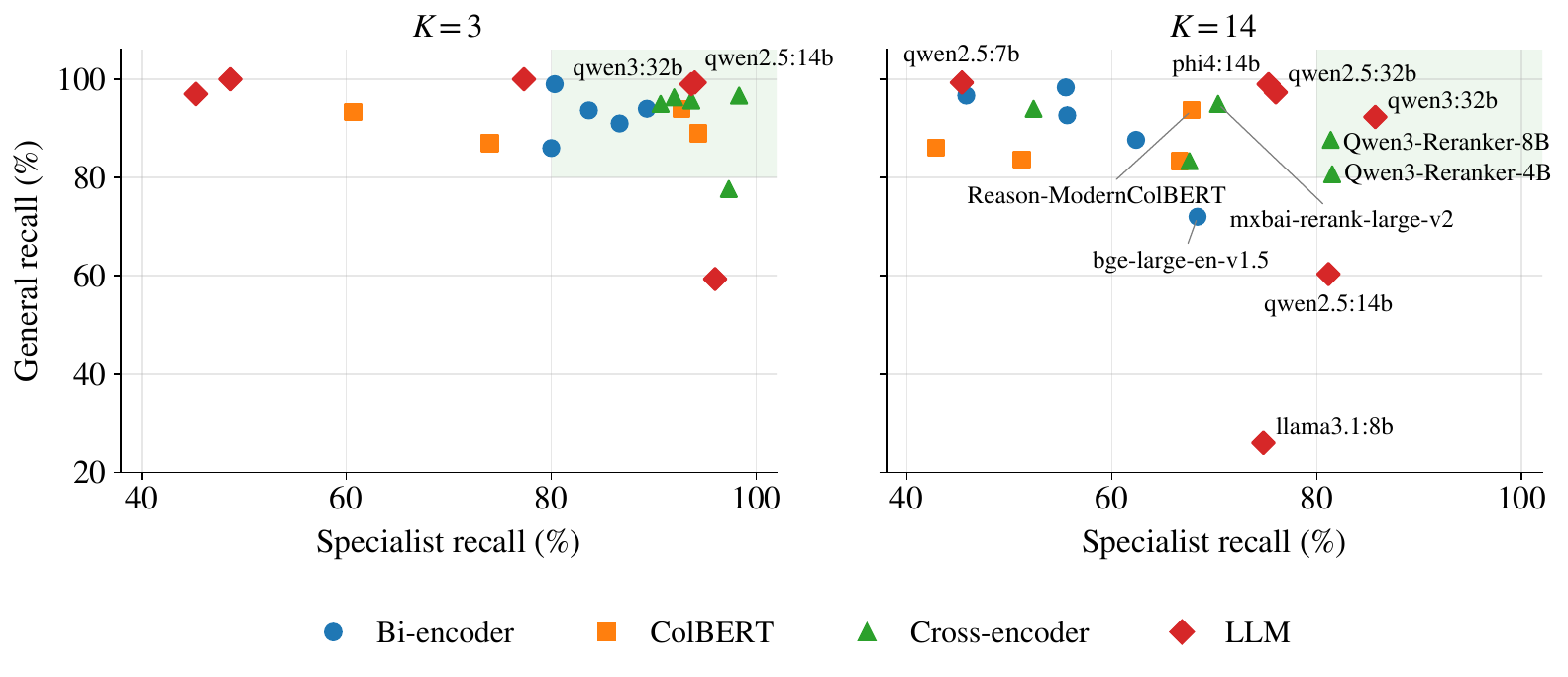}
\caption{Specialist recall (x-axis) and general recall (y-axis) of all 20 models at $K=3$ (left) and $K=14$ (right). 
The shaded zone marks both recalls above 80\%.}
\label{fig:recall}
\end{figure}

Two things stand out in Figure~\ref{fig:recall}. 
First, \textbf{retrieval routers route off-topic queries reliably, even when the pool is large}. 
At $K=14$, with the threshold $T(K)$ applied, every retrieval model keeps general recall above 70\%, whereas two of the six LLMs fall far below this level: qwen2.5:14b to about 60\% and llama3.1:8b to 26\%. 
Second, \textbf{the two best cross-encoders reach the same zone as the best LLM, with four to eight times fewer parameters}. 
At $K=14$ only three models have both recalls above 80\% (shaded zone): Qwen3-Reranker-8B, Qwen3-Reranker-4B and the LLM qwen3:32b. 
qwen3:32b uses thinking mode, so they generate a long reasoning text before they answer, whereas the cross-encoders only score the query against each description, yet they are still in the same zone. 
\noindent
The other retrieval models fall outside this region because their specialist recall declines as $K$ grows, to roughly 45\%-70\% at $K=14$. 
This is partly because it is harder to pick the right specialist among more candidates, and partly because $T(K)$ rises with $K$ to keep off-topic queries out, at the cost of sending a few borderline specialist queries to the general service.

\subsection{What does routing cost?}
\label{sec:res_cost}

\begin{figure}[H]
\centering
\includegraphics[width=\linewidth]{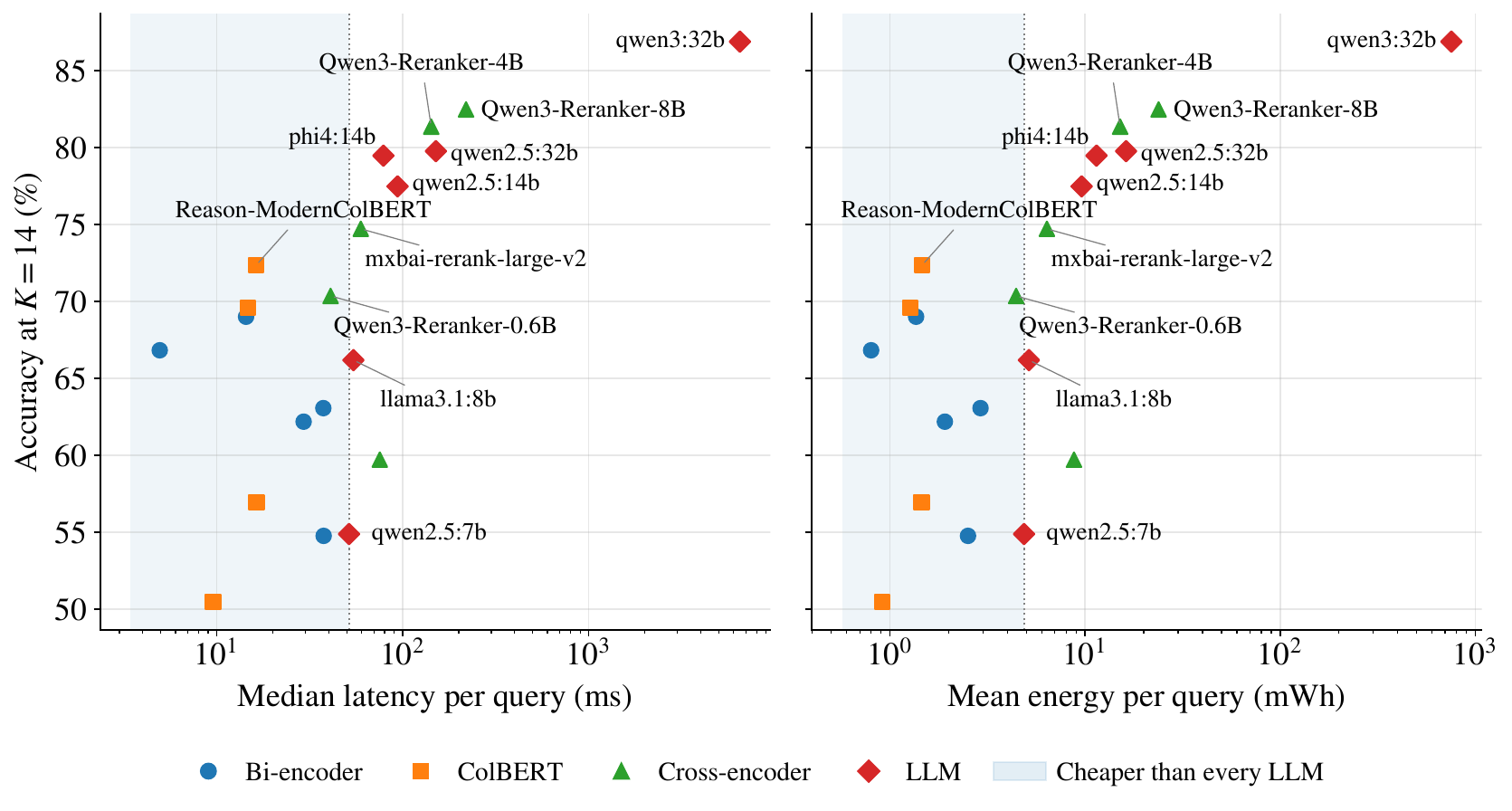}
\caption{Routing accuracy against median latency (left) and mean energy (right) per query at $K=14$ for all 20 models (log scale).}
\label{fig:cost}
\end{figure}

We examine $K=14$, the largest pool, and the dotted line marks the cheapest LLM, qwen2.5:7b (about 52\,ms and 5\,mWh per query). 
Figure~\ref{fig:cost} shows that \textbf{ten of the 14 retrieval models are both faster and more energy-efficient than every LLM, and five of these ten are also more accurate than the two smallest LLMs}.
The most noticeable ones are Reason-ModernColBERT and Qwen3-Reranker-0.6B, which are more accurate than the two smallest LLMs, llama3.1:8b and qwen2.5:7b (about 72\% and 70\%, against 66\% and 55\%).
They are also cheaper: Reason-ModernColBERT takes about 16\,ms and 1.5\,mWh per query and Qwen3-Reranker-0.6B about 41\,ms and 4.4\,mWh, against about 52 to 54\,ms and 5\,mWh for the two LLMs.
Moreover, \textbf{the most accurate retrieval models come close to the most accurate LLM at a fraction of its cost}.
Qwen3-Reranker-8B is about 4 points less accurate than qwen3:32b, but roughly 30 times faster and uses about 30 times less energy. 
Qwen3-Reranker-4B is about 5 points less accurate, but roughly 45 times faster and uses about 50 times less energy.
In the middle of the range, mxbai-rerank-large-v2 is cheaper than phi4:14b, qwen2.5:14b and qwen2.5:32b (about 60\,ms and 6\,mWh, against 80\,ms and 11\,mWh, 95\,ms and 10\,mWh, and 150\,ms and 16\,mWh), but about 3 to 5 points less accurate.

\begin{figure}[H]
\centering
\includegraphics[width=0.90\linewidth]{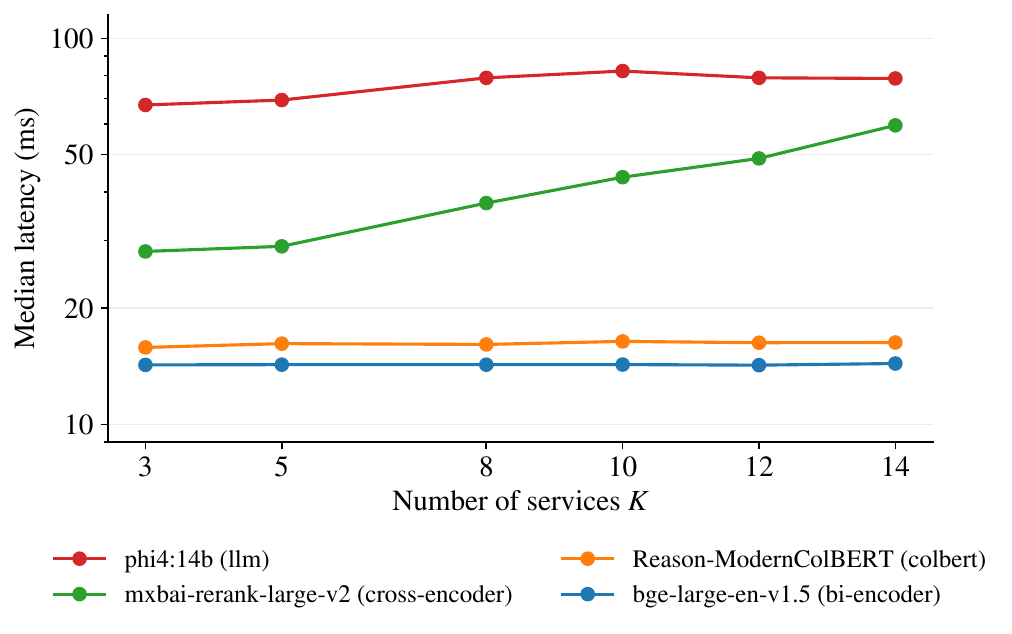}
\caption{Median latency per query as the number of services $K$ grows, for one model from each family: the LLM phi4:14b, the cross-encoder mxbai-rerank-large-v2, the ColBERT model Reason-ModernColBERT and the bi-encoder bge-large-en-v1.5.}
\label{fig:latency_k}
\end{figure}

Figure~\ref{fig:latency_k} shows that the bi-encoder and ColBERT models grow only minimally with $K$: bge-large-en-v1.5 stays at about 14\,ms and Reason-ModernColBERT at about 16\,ms. 
This is because the service descriptions are encoded once and reused for each $K$, so each query only needs to be encoded and compared with the stored descriptions.
The cross-encoder grows faster, because it reads every description together with the query: mxbai-rerank-large-v2 roughly doubles, from about 28\,ms to about 60\,ms.
The LLM phi4:14b grows slowly, from about 67\,ms to about 79\,ms, and stays above mxbai-rerank-large-v2 at every $K$.

\subsection{Does the gate work in a second domain?}
\label{sec:res_generalisation}

We applied the constants fitted on MMLU, without any refitting, to the MedMCQA setup described in Section~\ref{sec:MEDMCQA_setup}. 
Based on Table~\ref{tab:medmcqa_mmlu_constants}, every retrieval model sends most general queries to the general service, with general recall of roughly 60--96\% at $K=3$ and 74--99\% at $K=9$. 
So the gate keeps working as specialists are added, even in a new domain with new off-topic queries. 
The comparison with LLMs is also informative, since the strongest retrieval model, mxbai-rerank-large-v2, approaches several of them. 
It is the most accurate retrieval model at every $K$ (80.42\% at $K=3$, 73.33\% at $K=9$), matches qwen2.5:32b and beats llama3.1:8b at every $K$. 
It trails the strongest LLM overall, qwen3:32b, by about 4 to 11 points, at a fraction of its cost (Section~\ref{sec:res_cost}).
Specialist recall is relatively low for most models, including several LLMs, which is consistent with the overlap between medical specialties.
\begin{table}[H]
\centering
\caption{MedMCQA specialists with MMLU general queries (\%), using the constants fitted on MMLU without refitting.
Bold marks the strongest retrieval model.}
\label{tab:medmcqa_mmlu_constants}
\resizebox{\linewidth}{!}{%
\begin{tabular}{llrrrrrrrr}
\toprule
 & & \multicolumn{4}{c}{Acc.} & \multicolumn{2}{c}{Spec. Recall} & \multicolumn{2}{c}{Gen. Recall} \\
\cmidrule(lr){3-6}\cmidrule(lr){7-8}\cmidrule(lr){9-10}
 & & $K{=}3$ & $K{=}5$ & $K{=}7$ & $K{=}9$ & $K{=}3$ & $K{=}9$ & $K{=}3$ & $K{=}9$ \\
\midrule
Bi-encoder    & bge-large-en-v1.5        & 72.71 & 67.50 & 61.39 & 58.57 & 33.89 & 35.93 & 96.00 & 99.33 \\
              & Qwen3-Embedding-8B       & 65.83 & 65.17 & 61.81 & 61.43 & 47.22 & 42.78 & 77.00 & 95.00 \\
\midrule
ColBERT       & Reason-ModernColBERT     & 63.33 & 64.33 & 61.67 & 62.50 & 52.22 & 55.93 & 70.00 & 74.33 \\
              & LFM2-ColBERT-350M        & 72.92 & 70.67 & 63.33 & 62.50 & 58.33 & 54.26 & 81.67 & 77.33 \\
\midrule
Cross-encoder & mxbai-rerank-large-v2    & \textbf{80.42} & \textbf{80.00} & \textbf{74.17} & \textbf{73.33} & 65.56 & 62.78 & 89.33 & 92.33 \\
              & Qwen3-Reranker-4B        & 62.29 & 69.83 & 68.89 & 70.36 & 66.67 & 65.56 & 59.67 & 79.00 \\
\midrule
LLM           & phi4:14b                 & 78.12 & 83.00 & 77.64 & 77.74 & 54.44 & 66.48 & 92.33 & 98.00 \\
              & qwen2.5:14b              & 86.67 & 86.50 & 80.69 & 77.14 & 77.22 & 69.44 & 92.33 & 91.00 \\
              & qwen2.5:32b              & 80.62 & 80.00 & 75.14 & 72.50 & 48.89 & 57.22 & 99.67 & 100.00 \\
              & llama3.1:8b              & 57.08 & 64.50 & 66.67 & 65.95 & 79.44 & 67.41 & 43.67 & 63.33 \\
              & qwen3:32b               & 84.17 & 85.17 & 84.86 & 83.21 & 81.11 & 86.00 & 78.70 & 91.33
              \\
\bottomrule
\end{tabular}}
\end{table}

Refitting the constants on MedMCQA changes the results only slightly for most models.
Table~\ref{tab:transfer} shows that, for four of the six models, specialist recall changes by at most about 4 points and general recall by at most about 7 points. 
Only two models shift more: bge-large-en-v1.5 becomes stricter and loses about 13 points of specialist recall, and Qwen3-Reranker-4B becomes more permissive and lets about 6 to 9 points more general queries through. 
\begin{table}[H]
\centering
\caption{Specialist and general recall (\%) on MedMCQA with constants fitted on MedMCQA (Med.) and on MMLU (applied without refitting). $\Delta$ is MMLU minus Med., in points.}
\label{tab:transfer}
\resizebox{\linewidth}{!}{%
\begin{tabular}{lrrrrrrrrrrrr}
\toprule
 & \multicolumn{6}{c}{$K=3$} & \multicolumn{6}{c}{$K=9$} \\
\cmidrule(lr){2-7}\cmidrule(lr){8-13}
 & \multicolumn{3}{c}{Spec.} & \multicolumn{3}{c}{Gen.} & \multicolumn{3}{c}{Spec.} & \multicolumn{3}{c}{Gen.} \\
\cmidrule(lr){2-4}\cmidrule(lr){5-7}\cmidrule(lr){8-10}\cmidrule(lr){11-13}
Model & Med. & MMLU & $\Delta$ & Med. & MMLU & $\Delta$ & Med. & MMLU & $\Delta$ & Med. & MMLU & $\Delta$ \\
\midrule
bge-large-en-v1.5      & 47.78 & 33.89 & $-13.9$ & 87.00 & 96.00 & $+9.0$ & 48.33 & 35.93 & $-12.4$ & 91.00 & 99.33 & $+8.3$ \\
Qwen3-Embedding-8B     & 47.22 & 47.22 & $0.0$   & 77.00 & 77.00 & $0.0$  & 39.26 & 42.78 & $+3.5$  & 96.67 & 95.00 & $-1.7$ \\
Reason-ModernColBERT   & 52.22 & 52.22 & $0.0$   & 68.00 & 70.00 & $+2.0$ & 54.26 & 55.93 & $+1.7$  & 81.00 & 74.33 & $-6.7$ \\
LFM2-ColBERT-350M      & 58.89 & 58.33 & $-0.6$  & 81.33 & 81.67 & $+0.3$ & 55.19 & 54.26 & $-0.9$  & 70.67 & 77.33 & $+6.7$ \\
mxbai-rerank-large-v2  & 65.56 & 65.56 & $0.0$   & 89.33 & 89.33 & $0.0$  & 65.74 & 62.78 & $-3.0$  & 85.67 & 92.33 & $+6.7$ \\
Qwen3-Reranker-4B      & 63.33 & 66.67 & $+3.3$  & 68.67 & 59.67 & $-9.0$ & 63.15 & 65.56 & $+2.4$  & 85.33 & 79.00 & $-6.3$ \\
\bottomrule
\end{tabular}}
\end{table}

\section{Ablations}
\label{sec:ablations}

\subsection{Fixed versus $K$-aware threshold}
\label{sec:abl_threshold}

We compare the $K$-aware threshold with a \emph{fixed threshold}: the 90th percentile of the off-topic validation gaps at $K=3$, averaged over all groups of three specialists, and reused unchanged for every $K$.
Table~\ref{tab:ablation-threshold} illustrates that \textbf{a fixed threshold fails silently as specialists are added.} 
At $K=14$ it lets 34 to 68\% of off-topic queries reach a specialist, yet its accuracy is approximately 2 to 5 points below the $K$-aware threshold, because specialist queries make up most of the test set. 
Accuracy alone hides this failure, so we also compare general recall.
\textbf{The $K$-aware threshold fixes this at a small cost.} 
It sacrifices 1 to 5 points of specialist recall to gain 28 to 40 points of general recall, is more accurate for every model at every $K\geq 5$, and needs no retraining. 
The gain grows with $K$ and is largest at $K=10$, 12 and 14, which were not used for fitting.
These results are deterministic, since the retrieval models produce the same scores for the same input, so repeating the evaluation gave identical values.
 
\begin{table}[H]
\caption{Fixed versus $K$-aware threshold for each number of specialist services $K$. Acc., Spec.\ and Gen.\ are the overall accuracy, specialist recall and general recall (\%). 
The fixed threshold is calibrated once at $K=3$ and reused for every $K$. The $K$-aware constants are fitted on $K=3$ to $K=8$, so $K=10$, 12 and 14 are extrapolation. 
$\Delta$ is $K$-aware minus fixed, in points. Bold marks the higher accuracy.
\label{tab:ablation-threshold}}
\centering
\resizebox{\linewidth}{!}{%
\begin{tabular}{llrrrrrrrrrr}
\toprule
 &  &  & \multicolumn{3}{c}{\textbf{Fixed}} & \multicolumn{3}{c}{\textbf{$K$-aware}} & \multicolumn{3}{c}{\textbf{$\Delta$}} \\
\cmidrule(lr){4-6} \cmidrule(lr){7-9} \cmidrule(lr){10-12}
\textbf{Family} & \textbf{Router} & $K$ & \textbf{Acc.} & \textbf{Spec.} & \textbf{Gen.} & \textbf{Acc.} & \textbf{Spec.} & \textbf{Gen.} & \textbf{Acc.} & \textbf{Spec.} & \textbf{Gen.} \\
\midrule
Bi-encoder & bge-large-en-v1.5 & 3 & \textbf{91.83} & 89.33 & 94.33 & 91.67 & 89.33 & 94.00 & $-0.16$ & $0.00$ & $-0.33$ \\
 &  & 5 & 88.50 & 88.60 & 88.33 & \textbf{89.00} & 86.80 & 92.67 & $+0.50$ & $-1.80$ & $+4.34$ \\
 &  & 8 & 80.00 & 83.12 & 71.67 & \textbf{82.82} & 80.38 & 89.33 & $+2.82$ & $-2.74$ & $+17.66$ \\
 &  & 10 & 78.15 & 83.10 & 61.67 & \textbf{80.77} & 78.20 & 89.33 & $+2.62$ & $-4.90$ & $+27.66$ \\
 &  & 12 & 68.93 & 76.25 & 39.67 & \textbf{71.27} & 70.75 & 73.33 & $+2.34$ & $-5.50$ & $+33.66$ \\
 &  & 14 & 66.06 & 73.36 & 32.00 & \textbf{69.00} & 68.36 & 72.00 & $+2.94$ & $-5.00$ & $+40.00$ \\
\midrule
ColBERT & Reason-ModernColBERT & 3 & \textbf{91.83} & 94.33 & 89.33 & 91.67 & 94.33 & 89.00 & $-0.16$ & $0.00$ & $-0.33$ \\
 &  & 5 & 89.38 & 90.60 & 87.33 & \textbf{89.88} & 88.40 & 92.33 & $+0.50$ & $-2.20$ & $+5.00$ \\
 &  & 8 & 82.45 & 83.00 & 81.00 & \textbf{83.55} & 79.75 & 93.67 & $+1.10$ & $-3.25$ & $+12.67$ \\
 &  & 10 & 78.92 & 79.90 & 75.67 & \textbf{80.46} & 76.30 & 94.33 & $+1.54$ & $-3.60$ & $+18.66$ \\
 &  & 12 & 72.87 & 73.67 & 69.67 & \textbf{74.87} & 70.17 & 93.67 & $+2.00$ & $-3.50$ & $+24.00$ \\
 &  & 14 & 70.59 & 71.57 & 66.00 & \textbf{72.35} & 67.79 & 93.67 & $+1.76$ & $-3.78$ & $+27.67$ \\
\midrule
Cross-encoder & Qwen3-Reranker-4B & 3 & \textbf{97.83} & 98.33 & 97.33 & 97.50 & 98.33 & 96.67 & $-0.33$ & $0.00$ & $-0.66$ \\
 &  & 5 & 93.50 & 96.60 & 88.33 & \textbf{95.12} & 96.40 & 93.00 & $+1.62$ & $-0.20$ & $+4.67$ \\
 &  & 8 & 84.45 & 90.62 & 68.00 & \textbf{87.64} & 89.88 & 81.67 & $+3.19$ & $-0.74$ & $+13.67$ \\
 &  & 10 & 82.69 & 89.50 & 60.00 & \textbf{87.15} & 88.40 & 83.00 & $+4.46$ & $-1.10$ & $+23.00$ \\
 &  & 12 & 78.87 & 84.75 & 55.33 & \textbf{82.73} & 83.17 & 81.00 & $+3.86$ & $-1.58$ & $+25.67$ \\
 &  & 14 & 76.24 & 82.64 & 46.33 & \textbf{81.35} & 81.50 & 80.67 & $+5.11$ & $-1.14$ & $+34.34$ \\
\bottomrule
\end{tabular}}
\end{table}

\section{Discussion}
\label{sec:discussion}

The results indicate that reliable routing in modular AI systems is not solely a problem of identifying the most relevant specialist, but also of determining when no available specialist should be selected. 
This distinction becomes increasingly important as the specialist pool grows. 
While retrieval-based routing remains attractive because it avoids invoking multiple models to determine relevance, our experiments show that the statistical conditions under which the routing decision is made are not invariant to the number of available specialists. 
In particular, increasing $K$ changes the meaning of a given gap between the top and the median specialist score, so a fixed abstention threshold admits an increasing share of off-topic queries as the pool grows, whereas a threshold that accounts for $K$ keeps routing behaviour stable. 
More broadly, these findings suggest that scalability in modular AI is not determined only by computational cost or retrieval efficiency, but also by how reliably the router can preserve the distinction between \emph{``one specialist clearly matches''} and \emph{``none of the specialists sufficiently matches''} as the candidate pool expands.

One alternative to a threshold is to describe the general service like a specialist and treat it as one more service, so that a query goes to the general service only when its description scores highest. 
This removes the threshold, but it asks a single description to cover everything the specialists do not, such as banking requests, travel and small talk. 
Such a description must be very broad, and it would have to be rewritten whenever the off-topic queries change. 
A threshold avoids this, because it only asks whether any specialist fits well enough, not what the off-topic queries are about.

ADAPT can also be moved to a new domain with little effort. 
On MedMCQA, most models continued to reject the majority of off-topic queries using the constants fitted on MMLU, despite both the specialists and the off-topic queries originating from different sources. 
Where recalibration was required, it remained inexpensive, since the constants are estimated from off-topic queries alone and do not depend on labelled specialist queries. 
Deploying ADAPT in a new domain therefore requires only the specialist descriptions and a representative set of off-topic queries, with no additional model training.

Figure~\ref{fig:gap_boxplots} also suggests a scaling limit of the proposed approach. 
As $K$ increases, the gaps of specialist and off-topic queries move closer, and while the $K$-aware threshold compensates for the rise in off-topic gaps, it cannot separate distributions that increasingly overlap. 
Beyond some pool size, raising the threshold would reject more specialist queries, while lowering it would admit more off-topic queries, so the limit becomes one of discriminability rather than calibration. 
Our experiments up to $K=14$ show this trend but cannot determine where the limit occurs. 
This observation motivates hierarchical routing as a possible way to delay this limit. 
Rather than comparing a query against an increasingly large pool simultaneously, specialists could be organised into smaller groups, with an initial routing stage identifying a relevant group and a subsequent stage selecting among only its members. 
The abstention decision would then operate over a smaller effective candidate pool, where the separation between specialist and off-topic queries may remain greater. 
Such a design could preserve the computational advantages of retrieval-based routing while allowing substantially larger collections of specialists, although this requires further investigation.

\section{Limitations and Future Work}
\label{sec:limitation}

The main limitation is that we evaluated pools of up to 14 specialists on MMLU and 9 on MedMCQA. 
As $K$ grows, off-topic gaps move closer to in-scope gaps, so for much larger pools a single threshold may no longer separate the two. 
Section~\ref{sec:discussion} discusses this limit and how hierarchical routing could extend ADAPT to larger pools. 
Moreover, the constants do not always transfer to a new domain, since on MedMCQA most models could reuse the MMLU constants but a few needed refitting (Section~\ref{sec:res_generalisation}). 
Additionally, we modelled the threshold as a function of $\ln K$, which fits our range well, but other forms, such as linear in $K$ or in $\sqrt{K}$, were not explored. 
Furthermore, our study focuses on English text queries with a single intent, while other languages, multi-intent queries and other modalities such as images and audio lie outside its scope.
Finally, we tested each LLM with a single zero-shot prompt (Appendix~\ref{app:router_prompt}), the same for all six models, which asks for the service name only, without examples or reasoning. 
Other prompts may change their accuracy, although longer prompts would also make each query slower and more costly to route.

In our next work, we will compare these functional forms, test ADAPT on much larger pools, study whether it can be applied to new domains without refitting, and implement and evaluate hierarchical routing in a real modular AI platform such as CoSMIC.

\section{Conclusions}
\label{sec:conclusions}

This paper examined whether retrieval models can act as practical service routers that either select an appropriate specialist or abstain to a general service when no suitable specialist exists, and found that they can, provided the abstention threshold grows with the number of specialists.
The central observation is that off-topic queries behave differently as a system grows. 
Their gaps rise steadily with pool size.
This movement is consistent across bi-encoders, ColBERT models and cross-encoders, and it follows $\ln K$ closely. 
A fixed threshold ignores it and therefore fails, letting in 34 to 68\% of off-topic queries at 14 specialists.
A threshold that moves with it, $T(K)$, raises general recall by 28 to 40 points for only 1 to 5 points of specialist recall.
This gain holds at pool sizes not used for fitting, and the threshold keeps rejecting most off-topic queries in a second domain without refitting.
With this gate, retrieval models remain competitive with LLMs at a fraction of their cost.
A router's threshold should therefore not be a constant, but follow how off-topic queries behave as the pool grows.

\authorcontributions{Conceptualization, M.S. and M.A. and C.C.N.K.; methodology, M.S. and M.A. and C.C.N.K.; software, M.S. and M.A.; validation, M.S. and M.A. and A.S. and C.C.N.K.; formal analysis, M.S. and M.A.;
investigation, M.S.; writing, M.S. and M.A. and A.S. and C.C.N.K.; supervision, C.C.N.K. All authors have read and agreed to the published version of the manuscript.}
\funding{This work is funded under the agreement with the ACT Government, Future Jobs Fund – Open Source Institute (OpenSI) – R01553; and NetApp Technology Alliance Agreement with OpenSI – R01657. 
This research was supported by the Australian Government through the Department of Education’s National Industry PhD Program (project 36337). 
The views expressed herein are those of the authors and are not necessarily those of the Australian Government or the Department of Education.}
\dataavailability{Our code will be made available upon publication.}
\acknowledgments{During the preparation of this manuscript, the authors used Claude Sonnet 5.5 and GitHub Copilot with GPT-5.6 for debugging, assisting in the preparation of figures and refining the draft text. 
The authors have reviewed and edited all AI-assisted output and take full responsibility for the content of this publication.}
\conflictsofinterest{The authors declare no conflicts of interest.}

\newpage
\appendixtitles{yes}
\appendixstart
\appendix

\section{Percentile Sensitivity Analysis}
\label{app:percentile}

We chose the percentile $p$ on a validation set with specialist queries from MMLU and general queries from BANKING77 and CLINC-OOS.
For $K = 3, 5, 8$ and 10, it contains 41, 63, 118 and 151 specialist queries, respectively, and 50 general queries each.
For each model and $p$, we applied $T(K)=c_0+c_1\ln K$ with the constants fitted at that percentile (Table~\ref{tab:percentile}).

\begin{table}[H]
\caption{Validation results for each percentile $p$, averaged over 14 models and $K = 3, 5, 8, 10$. Balanced accuracy is the mean of specialist and general recall.}
\label{tab:percentile}
\centering
\begin{tabular}{lrrr}
\toprule
$p$ & Specialist recall & General recall & Balanced accuracy \\
\midrule
85 & 0.868 & 0.863 & 0.865 \\
90 & 0.852 & 0.921 & 0.886 \\
95 & 0.821 & 0.968 & 0.894 \\
99 & 0.748 & 0.991 & 0.870 \\
\bottomrule
\end{tabular}
\end{table}

A higher $p$ makes the gate stricter, raising general recall and lowering specialist recall. 
Balanced accuracy is highest at $p=95$, but $p=90$ is within 0.01 of it and rejects fewer specialist queries (0.852 against 0.821). 
We therefore use $p=90$ for all models and all $K$, because incorrectly rejecting a specialist query loses an answer the user asked for.

\section{Prompts for Description Construction}
\label{app:prompts}

\subsection{Concept Extraction}
\begin{lstlisting}
You are analyzing validation questions from an academic subject.

SUBJECT:
{subject}

VALIDATION QUESTIONS:
{formatted_questions}

TASK:
Analyze all of the questions and answer choices above.

Extract the domain knowledge concepts that the questions are testing.

Use the answer choices only as context for understanding what knowledge
the question tests. Do NOT treat every answer choice or named entity as
a separate concept.

For extracted_concepts:
- Identify the specific domain knowledge concept, principle, structure,
  mechanism, theory, or process being tested by each question.
- Prefer meaningful and reusable domain concepts over isolated entities,
  answer options, or question-specific wording.
- Use the narrowest concept that accurately represents the knowledge
  being tested, while avoiding unnecessary details that apply only to
  one particular question.
- Merge concepts only when they are true semantic near-duplicates and
  represent essentially the same knowledge.
- Keep concepts separate when they represent genuinely different
  knowledge, even if they belong to the same broader topic.
- Ensure the extracted concept set collectively covers the full
  validation set.
- Multiple questions may map to the same concept when they genuinely
  test closely related knowledge.
- Do NOT force one concept per question.
- Exclude concepts that appear only as irrelevant distractor choices.
- Do not simply copy answer choices.
- Do not introduce concepts unsupported by the validation questions.
- Do not broaden specific concepts into vague umbrella categories.
- Use concise, natural-language noun phrases.
- Do not use snake_case.
- Do not answer the questions.
- There is no required number of extracted concepts.

Return ONLY valid JSON in this format:

{{
    "subject": "{subject}",
    "extracted_concepts": []
}}
\end{lstlisting}

\subsection{Core Concept Selection}
\begin{lstlisting}
You are constructing a semantic profile for an academic routing service.

SUBJECT:
{subject}

EXTRACTED CONCEPTS:
{concepts_text}

TASK:
Select the core concepts that best characterize the knowledge domain
represented by this subject.

For core_concepts:
- Select 6-9 concepts when at least 6 valid concepts are available.
- Select only from the extracted concepts provided above.
- Do not introduce new concepts.
- Prefer concepts that are central, specific, and representative of the
  subject's knowledge domain.
- Prefer concepts that are specific and characteristic of this subject,
  rather than concepts that could broadly describe many academic domains.
- Avoid vague umbrella concepts.
- Avoid redundant or near-duplicate concepts.
- Select concepts from different important areas of the subject when possible.
  Avoid selecting several concepts that represent essentially the same subarea
  while excluding other clearly represented subareas.
- Do not broaden concepts beyond their original meaning.
- If fewer than 6 extracted concepts are available, retain all valid concepts.
- There is no need to select 9 concepts if fewer concepts adequately
  represent the domain.

Return ONLY valid JSON in this format:

{{
    "subject": "{subject}",
    "core_concepts": []
}}
\end{lstlisting}

\subsection{Specialist Description}
\begin{lstlisting}
You are constructing a semantic description for an academic specialist
routing service.

SUBJECT:
{subject}

CORE CONCEPTS:
{core_text}

FULL EXTRACTED CONCEPTS:
{extracted_text}

TASK:
Create one concise semantic description of the academic knowledge covered
by this specialist service.

The core concepts represent the strongest anchors of the subject.
The full extracted concepts provide broader coverage of the knowledge
represented by the subject.

For the description:
- Use the core concepts as the primary anchors.
- Use the full extracted concepts to preserve broader subject coverage.
- Represent multiple important areas of the subject when supported by
  the concepts.
- Summarize related concepts into meaningful domain areas rather than
  listing every concept individually.
- Begin directly with the subject name or subject-specific knowledge.
- Avoid generic boilerplate openings such as "This domain encompasses",
  "This service covers", or "The academic knowledge domain encompasses".
- Prioritize domain-specific terminology over generic descriptive language.
- Describe the academic knowledge domain rather than individual questions.
- Use only information supported by the supplied concepts.
- Do not introduce unsupported topics.
- Do not mention validation questions, concept extraction, routing
  experiments, or benchmarking.
- Do not use procedural wording such as "choose this service when".
- Use concise, natural academic language.
- Write between 40 and 60 words.

Before returning the final description:
- Refine the description for clarity, coherence, and semantic precision.
- Remove unnecessary repetition and generic boilerplate language.
- Ensure the wording is natural and academically appropriate.
- Ensure the subject name is written naturally rather than in snake_case.
- Check that the description represents both the core concepts and the
  broader extracted concepts.
- Check that no unsupported concepts have been introduced.

Return ONLY valid JSON in this format:

{{
    "subject": "{subject}",
    "description": ""
}}
\end{lstlisting}

\subsection{Example description of a specialist}
\begin{quote}
Abstract algebra focusing on group theory structures, including element and subgroup orders, Lagrange's theorem, homomorphisms, normal subgroups, factor groups, and isomorphism theorems. Encompasses symmetric groups, permutation properties, integral domains, fields of quotients, and applications of Euler's phi function in structural analyses.
\end{quote}

\section{Prompt for LLM Routers}
\label{app:router_prompt}

\begin{lstlisting}
You are a query router for a multi-service assistant system. You will be
given a list of available specialist services with short descriptions,
followed by a user query. Decide which single specialist service is best
suited to handle that query, using only the descriptions provided below.

Rules:
- Match the query against the core intent of each specialist service
  description, not just surface keywords.
- If more than one specialist service seems plausible, choose the one whose
  description is the more specific and direct match.
- If the query does not match any of the listed specialist services,
  return -1.
- For a fallback query, respond exactly 'service -1'.
- Never assume a specialist service exists beyond those listed below.
- Respond only in the exact form 'service $name'.
- For a specialist query, $name must exactly match one of the listed
  specialist names.
- Do not include any explanation or additional text.

Specialist services:
{active_services_joined}

Query: "{query}"

Which service should handle this query?
\end{lstlisting}

\newpage

\reftitle{References}
\bibliography{references}

\end{document}